\documentclass[journal,10pt, onecolumn]{IEEEtran}

\newtheorem{thm}{Theorem}
\newtheorem{remk}{Remark}

\newtheorem{lem}{Lemma}

\usepackage[final]{graphicx}
\usepackage[reqno]{amsmath}
\usepackage{amssymb}
\usepackage{subfig}
\usepackage{bm}
\usepackage{bbm}
\usepackage{subfig}
\usepackage{algorithm}
\usepackage{algpseudocode}
\usepackage{flushend}

\begin{document}

\sloppy
\title{Multi-user Multi-armed Bandits for Uncoordinated Spectrum Access\thanks{Parts of this work were presented at  ICNC \cite{bande19icnc} and submitted to ICASSP.}}

\author{
        Meghana~Bande,~\IEEEmembership{Student~Member,~IEEE,}
        and~Venugopal~V.~Veeravalli,~\IEEEmembership{Fellow,~IEEE}
\thanks{M. Bande and V. V. Veeravalli are with the Coordinated Science Laboratory and the Department of Electrical and Computer Engineering, University of Illinois
at Urbana-Champaign, Urbana, IL 61801 USA (e-mail: mbande2@illinois.edu, vvv@illinois.edu).}
\thanks{This research was supported by the US NSF WIFiUS Program under grant number CNS 14-57168 and by the US NSF SpecEES under grant number 1730882, through the University of Illinois at Urbana-Champaign.}}
\maketitle

\begin{abstract}
A multi-user multi-armed bandit (MAB) framework is used to develop algorithms for uncoordinated spectrum access. The number of users is assumed to be unknown to each user.  A stochastic setting is first considered, where the rewards on a channel are the same for each user. 
In contrast to prior work, it is assumed that the number of users can possibly exceed the number of channels, and that rewards can be non-zero even under collisions. The proposed algorithm consists of an estimation phase and an allocation phase. It is shown that if every user adopts the algorithm, the system wide regret is constant with time with high probability. The regret guarantees hold for any number of users and channels, in particular, even when the number of users is less than the number of channels. Next, an adversarial multi-user MAB framework is considered, where the rewards on the channels are user-dependent. It is assumed that the number of users is less than the number of channels, and that the users receive zero reward on collision. 
The proposed algorithm combines the Exp3.P algorithm developed in prior work for single user adversarial bandits with a collision resolution mechanism to achieve sub-linear regret. It is shown that if every user employs the proposed algorithm, the system wide regret is of the order $O(T^\frac{3}{4})$ over a horizon of time $T$. The algorithms in both stochastic and adversarial scenarios are extended to the dynamic case where the number of users in the system evolves over time and are shown to lead to sub-linear regret.

 \end{abstract}


\section{Introduction}

The existing spectrum management paradigm treats frequency spectrum as a fixed commodity, which leads to spectrum under-utilization. Cognitive radio has emerged as a useful strategy to increase spectrum utilization. The existing literature on cognitive radio has largely been focused on the primary/secondary user paradigm, where secondary users need to detect vacant spectrum when available and vacate the occupied spectrum when a primary user wants to transmit. 

We focus on a different type of spectrum sharing system in which there is no distinction between users, and in which there is no coordination among the users. The collective performance across all users
is more important than that of individual users. This is in contrast to the typical primary/secondary user
paradigm in which secondary users bear the responsibility for ensuring priority-based spectrum sharing.
We model this system using an adversarial multi-user multi-armed bandit (MAB) framework \cite{bubeck12survey}. Our goal is to design an efficient channel access mechanism by managing interference in the system through a decentralized policy across the users.


Multi-arm bandit formulations in stochastic multi-user cognitive radios without user coordination were considered in \cite{Zhao10distributed}, \cite{Anand11}, \cite{AvnerM14mega} and \cite {RosenskiSS15mc}. The algorithm  in  \cite{Zhao10distributed} is based on a time-division fair sharing (TDFS) of the best arms between users. Although the algorithm achieves order optimal regret asymptotically, it requires pre-agreement among users and it is assumed  that the number of users is fixed and known to all users. The algorithm in \cite{Anand11} does not require any coordination between users and achieves optimal regret asymptotically, but assumes that the number of users is known. The algorithm in \cite{AvnerM14mega} combines an $\epsilon$-greedy learning rule with a collision avoidance mechanism, and \cite {RosenskiSS15mc} considers a musical chairs algorithm. Both of these approaches achieve sub-linear regret and do not require knowledge of the number of users. 
However, it was assumed that the channel parameters are the same for all the users.  A stochastic multi-user MAB with user dependent rewards on channel was considered in \cite{kalathil2014decentralized}.  However, the algorithm considers coordination and communication between users via an auction algorithm.

In this work, we focus on two scenarios that have not been previously studied in the multi-user MAB setting for uncoordinated dynamic spectrum access. We assume that the number of users is unknown and that there is no communication between the users.  However, we make the mild assumption that the users have access to a  shared clock for time synchronization (see also, \cite{RosenskiSS15mc,nieminen2009time,avner2015learning}).

We first study a stochastic multi-user MAB where the rewards on the channels are not user dependent. In our model, all users are treated equally and the reward obtained by each user largely depends on the actions of the other users. When multiple users access the same channel, we allow for a non-zero reward with the assumption that the reward for each user decreases as the number of users on the channel increases. Thus we include the case where there are more users than channels. 
This is in contrast to the existing approaches, including \cite{AvnerM14mega} and \cite{RosenskiSS15mc}, which focus on the primary/secondary user paradigm in the scenario where the reward distribution for a user is unknown but fixed. In particular, when multiple users access the same channel they receive zero reward. Hence, all these approaches fail when the number of users is greater than the number of channels.  

We assume that the reward on the channel depends on the number of users on the channel and is drawn i.i.d from a distribution depending on the number of users on the channel. The degradation of the reward as a function of number of users depends on the system, e.g., the distance between the users, the protocol used for transmission (e.g., hybrid ARQ) and is captured through a reward distribution that depends on the number of users on the channel.

 We propose an algorithm and show that if each user employs the algorithm, the system wide regret is $O(1)$ in time, with high probability. The algorithm can be used for any number of users or channels. 
 To the best of our knowledge, we are the first to provide sub-linear regret guarantees without user coordination when the number of users is greater than the number of channels.

In the second scenario, we study the adversarial multi-user MAB framework with user-dependent rewards. The adversarial bandit problem is an important variation of the MAB problem, where no stochastic assumption is made on the generation of rewards. The term ``adversarial" refers to the mechanism choosing the sequence of rewards on each arm.  If this mechanism is independent of the userÕs actions, then the adversary is said to be \emph{oblivious}. If the mechanism may adapt to the users' past behaviors, then the adversary is said to be \emph{non-oblivious} \cite{bubeck12survey}. The existing literature on adversarial MABs is focused on the single user case, and a detailed overview of the proposed solutions for the adversarial MAB formulation can be found in \cite{bubeck12survey}. The proposed algorithms in the single user adversarial setting achieve a sub-linear regret of $O(\sqrt{T})$ over a time horizon $T$. 

 We consider multi-user dynamic spectrum allocation without any coordination among the users. We also assume that the rewards on each channel are user-dependent and may vary with time. Such a system is captured through a multi-user adversarial MAB model, particularly when the reward distribution for each channel and user may change over time. We propose an algorithm, and show that if each user employs the algorithm, the system wide regret is $O(T^{\frac{3}{4}})$ over a time horizon $T$. To the best of our knowledge, we are the first to consider the multi-user setting for adversarial MABs and to provide sub-linear regret guarantees.  

\section{System Model and notation}

Let $K$ be the number of users in the system. We initially assume that the users have unlimited data for transmission. In a more realistic setting, users may become active or inactive depending on their transmission needs; our dynamic setting covers this scenario. Each user can choose one among $M$
channels for transmission. With $M$ channels and $K$ users attempting to access the spectrum, we assume that each user has prior knowledge of $M$, but not of $K$. The assumption of known $M$ is reasonable if the spectrum partition is enforced and fixed. On the other hand, it is not realistic to assume the knowledge of $K$ in an uncoordinated network.

We model the system as a multi-user MAB system with $K$ users and $M$ arms (channels). 
In each time unit $t$, let ${\cal A}^{k}_{t}$ denote the set of channels available to user
$k$. User $k$ chooses a channel $a^k_{t}\in {\cal A}^{k}_{t}$  based on the reward history according to a certain policy and receives a reward $g^k_{t}$. 
We assume that $g^k_{t}\in [0,1]$, and that each user chooses a channel according to the same algorithm. The reward on each arm depends on the number of users who have chosen
the arm. Let $f_t = [f_t(1), \ldots,f_t(M)]$ denote the number of users on each channel at time $t$, where $\sum_{m=1}^{M}f_t(m) = K$. Thus, the reward $g^k_{t}(a^k_t,f_t(a^k_t))$ received by user $k$  at time $t$ is a function of the channel chosen $a^k_t$ and the number of users on the channel $f_t(a^k_t)$. 

\subsection{Stochastic setting}

We model the system as a stochastic multi-user MAB system with $K$ users and $M$ arms (channels).  Each user can choose one among $M$
channels for transmission, where we allow for the possibility that $K\geq M$. 
We assume that the reward observed is inversely proportional to the number of users transmitting
on the same channel. For example, the reward could be the rate achieved by the user on the channel which reduces due to interference from other users accessing the channel.
Let $\mu(m,f(m))$ denote the mean reward on channel $m$ when the number of users on the channel is $f(m)$. We assume that each user chooses a channel according to the same policy.
We assume that $\mu(m,f(m))$ becomes negligible for some $f(m)=\beta+1$, where $\beta$ depends on the system. This restricts the number of users in the system as $\frac{K}{M}\leq \beta$.

In order to ensure that one user does not monopolize a channel for an extended period of time, we impose the following condition. For each user, transmission on a particular channel takes place for a maximum of $T_x$ time units, after which the user releases the channel for at least $T_x$ time units before attempting to access the same channel. 

We define the expected regret in the system as
\vspace*{-0.1 in}
\[
\mathbb{E}[{R(T)}] = T\sum_{i=1}^{M}f^{\ast}(i)\mu(i,f^{\ast}(i)) - \sum_{t,k}\mathbb{E}[g^k_{t}(a^k_t,f_t(a^k_t))]
\]
where $f^{\ast}= \text{argmax}_{f}\sum_{i=1}^{M}f(i)\mu(i,f(i))$ corresponds to the optimal number of users on each channel.

To estimate the means on each channel as a function of number of users, we need to impose the following separability condition.

For any $m\in[M]$ and $r,s\in [\beta]$ and some $\epsilon_2\in(0,1)$,
\begin{equation}\label{eqn:sep}|\mu(m,r)-\mu(m,s)|\geq 4Mc\exp \left(\frac{K-1}{M-1}\right)\sqrt{\sigma^{2}+\epsilon_{2}},\end{equation}
where
$\sigma^2$ is the variance of the distributions and $c$ is a constant.

\subsection{Adversarial setting}

In this case, we model the system as an adversarial multi-user MAB with $K$ users and $M$ channels. 
 We further restrict attention to the setting where there are more channels than users in the system i.e., $K\leq M$. 
We assume that each user chooses a channel according to the same algorithm. For user $k\in [K]$, let $p^k_{t} = (p^k_{t+1}(1),...,p^k_{t+1}(M))$ denote the probability vector across the arms, where $p^k_{t}(m)$ is the probability of choosing arm $m$ at time $t$.
We assume that the adversary chooses different reward for different users for the same channel. Let $g^k_{t}(a^k_t, f(a^k_t))$ denote the reward observed by user $k$ on choosing channel $a^k_t$ at time $t$. We assume that if more than one user chooses the same channel, they all receive zero reward. In other words, the users observe zero reward on collision. If there is no collision on the channel, the user observes a reward that is chosen by an adversary. 
Thus, we set $g^k_{t}(a^k_t)=0$ when $f(a^k_t)> 1$. 

We adopt the standard notion of pseudo-regret used for adversarial bandits in \cite{bubeck12survey}. The expected total regret in the system until time $T$ is defined as 
\[\mathbb{E}[R(T)] = \max_{{\cal K}:{\cal K}\subseteq [M],|{\cal K}|=K}\mathbb{E}\left[\sum_{t=1}^{T}\sum_{i\in {\cal K}}g^{k}_{t}(i) - \sum_{t=1}^{T}\sum_{k=1}^{K}  g^{k}_{t}(a^k_t)\right].\]

\section{Stochastic setting}

In this section, we focus on the stochastic multi-user MAB with user-independent rewards on each channel. We present an algorithm which leads to sub-linear regret with high probability, and extend it to the dynamic case.

\subsection{Algorithm}\label{sto:alg}

The algorithm has two phases. The first is an estimation phase during which we estimate the number of users $K$ and $\mu(m,f(m))$, the average mean reward on each channel as a function of the number of users on the channel. The second is an allocation phase where the users arrange themselves in a way that minimizes system regret.

\begin{algorithm}[h]
  \caption{}
  \label{sto:main}
  \begin{algorithmic}[1]
 \For {$t=0 \text{ to } T_0$}
\State $m\sim U(M)$
\If {no collision}
\State co$_m \gets$ co$_m+1$
\State $x_1(m)\gets x_1(m)+r(t)$
\Else
\State append $r(t)$ to $x(m)$
\State $\eta_c\gets\eta_c+1$
\EndIf
  \EndFor
  \State $\hat{K}\gets \min \{1+\text{round}\left(\frac{\ln (\frac{T_0-\eta_c}{T_0})}{\ln(1-\frac{1}{M})}\right),\beta M$\} and $\hat{\mu}(:,1)\gets \frac{x_1}{\text{co}}$
  \If {$\hat{K}> M$}
  \State   $\hat{\mu}(m,2:\beta)\gets$  Cluster $(x(m))$ for all $m$
  \State Calculate $f^{\ast}$ from $\hat{\mu}(m,f),\hat{K}$
\State Permute($N_0,T_f+T_x,\infty$)
\Else
\State ch =  Alloc($\hat{M},T_f+T_x$) where $\hat{M}$ is set of $\hat{K}$ best channels
\State After $T_x$, choose ch+1 in  $\hat{M}$ for next $T_x$ time units
\EndIf
  \end{algorithmic}
\end{algorithm}

We estimate the number of users by keeping track of the number of collisions similar to  \cite{RosenskiSS15mc}, with the 
estimate given by  $\hat{K}=\min \{1+\text{round}\left(\frac{\ln (\frac{T_0-\eta_c}{T_0})}{\ln(1-\frac{1}{M})}\right),\beta M$\}.

 
\begin{algorithm}[h]
  \caption{Cluster}
  \label{sto:cluster}
  \begin{algorithmic}[1]
 \State Run an $\alpha$-approximation algorithm for the k-means problem on input $X$, obtain $\beta$ means $\nu_1,\dots,\nu_{\beta}$
 \State $S_r \gets \{i: |x_i-\nu_r |\leq  |x_i-\nu_s | \text{ for every } $s$ \}$
 \State Return $g(S_r)=\frac{1}{|S_r|}\sum_{i\in S_r}x_i$
  \end{algorithmic}
\end{algorithm}

We estimate $\mu(m,n)$ separately for each channel based on the reward $x(m)$ observed on the corresponding channel, by clustering the samples using the k-means algorithm. We employ the algorithm Cluster (see Algorithm \ref{sto:cluster}) inspired by \cite{tang16}. We are interested in finding the centroids of the clusters rather than the correct classification of all the samples. Hence, we use an $\alpha$-approximation algorithm with a run time $T_c$ to find the estimates the centroids of the cluster and show that we get good estimates with high probability. We consider the approximation algorithm in \cite{kumar2004approx} with a run time $T_c\sim O(T_0)$.

\begin{algorithm}[h]
  \caption{Alloc}
  \label{sto:alloc}
  \begin{algorithmic}[1]
 \For {$t=1$ to $T$}
 \State $a_{t}\sim U(A)$
 \If {$\mu(a_t,f(a_t))\geq \mu(a_t,f^{\ast}(a_t))$}
 \State Choose action $a_{\tau}=a_t, \quad \forall \tau\geq t$  
   \EndIf
  \EndFor
  \end{algorithmic}
\end{algorithm}

\begin{algorithm}[h]
  \caption{Permute}
    \label{sto:permute}
  \begin{algorithmic}[1]
\State $A_1 = [M]$
  \For {$i = 1 \text{ to } N_0$} 
  \State $q(i)$ = Alloc($A_i,T_f+T_x$); 
  \State  $A_i \gets [M]\backslash \{q(i)\}$
 \EndFor
 \While{$t \leq T_1$}
      \State $j = t\mod N_0$
  \State Choose $q(j)$ for next $\min\{T_1,j(T_x+1)-1\}$ rounds 
  \EndWhile
  \end{algorithmic}
  \label{alg:permute}  
\end{algorithm}

After obtaining estimates for $\hat{\mu}(m,f)$ and $\hat{K}$, the optimal number of users on each channel $f^{\ast}$ can be calculated. We use Alloc (see Algorithm \ref{sto:alloc}) to ensure that each user settles or `fixes' on a channel $m$ , for which the number of users less than $f^{{\ast}}(m)$. That is, on finding a channel $m$ with $\mu(m,f(m))\leq \mu(m,f^{\ast}(m))$, the user keeps transmitting on it for at most $T_x$ time units. The system incurs regret until all users have settled on some channels,  and we call this duration the \emph{fixing time}. Once all the users have settled on their channels the system does not incur regret. However in our system model, a user can transmit on a channel for at most $T_x$ time units, after which the user must switch. We assume that $T_x$ is fixed for all the users but can vary with time. 
We use Permute (see Algorithm \ref{sto:permute}) to construct an  efficient allocation for which the regret does not grow with time. 
In order to avoid system-wide regret every time users have to switch, we fix the ordering of each user after $N_0$ epochs; this can be done for any $N_0\geq 2$. 
Our goal is to have each user transmit on all the channels. This is the coupon collector problem with each user having to collect $M$ channels with the expected number of trials $N_0 \sim  O(M\text{log} M)$. 
When $K\leq M$, in order to have efficient allocation so that the regret does not grow with time, after the first epoch, each user switches to the next channel among the set of $K$ best channels.

We fix the epoch size to be $T_x+T_f$, where $T_f$ is the expected time taken for all the users to fix on a channel. After $N_0$ epochs, we continue with an epoch size of $T_x$. We assume that 2 $\max_{m}f^{{\ast}}(m)\leq \sum_{m}f^{{\ast}}(m)$ to ensure that after every transmitting for $T_x$ time units, each user has other available channels. Note that our algorithm works even when $K\leq M$, in which case it reduces to a version of the algorithm in \cite{RosenskiSS15mc}.

\subsection{Analysis}\label{sto:analysis}

We first investigate the case where $K > M$. We show that if all the users in the system use Algorithm \ref{sto:main}, with high probability, the expected regret is $O(1)$ in $T$.

\subsubsection{Estimation phase}

We  now show that, with high probability, we have the correct estimates for $\mu(m,f(m))$. More precisely, we find estimates $\hat{\mu}^k(m,n)$ such that $|\hat{\mu}^k(m,n)-\mu(m,n)|\leq \epsilon$ with high probability.

\begin{lem}\label{sto:mu}
For any fixed $\epsilon, \delta$, user $k$, channel $m$ and number of users on the channel $n\leq \beta$ the estimate $\hat{\mu}^k(m,n)$ obtained after running the algorithm for $T_0 = \left\lceil  \frac{32\exp(\frac{K-1}{M-1})M}{\epsilon^2}\ln \frac{2MK\beta(\beta+1)}{\delta} \right\rceil$, and the $\alpha$ approximation algorithm for $T_c\sim O(T_0)$ rounds, we have with probability at least $1-\delta$,
\[|\hat{\mu}^k(m,n)-\mu(m,n)|\leq \epsilon.\]
\end{lem}

\begin{IEEEproof}
Let $A_1$ denote the event that there is at least one combination $k,m,n$ such that $|\hat{\mu}^k(m,n)-\mu(m,n)|\geq \epsilon$ and $A_2$ denote the event that each player has more than $\frac{16}{\epsilon^2}\ln \frac{2MK\beta(\beta+1)}{\delta}$ observations from distribution with mean $\mu(m,n)$ for each $m,n$.
\begin{eqnarray*}
\Pr(A_{1}) & = & \Pr(A_{1}|A_{2})\Pr(A_{2})+\Pr(A_{1}|A_{2}^{c})\Pr(A_{2}^{c})\\
 & \leq & \Pr(A_{1}|A_{2})+\Pr(A_{2}^{c}).
\end{eqnarray*}
It suffices to show that $\Pr(A_{1}|A_{2})\leq \frac{\delta}{2}$ and $\Pr(A_{2}^{c})\leq \frac{\delta}{2}$. From Lemma \ref{thm:num} in the appendix, we have $\Pr(A_{2}^{c})\leq \frac{\delta}{2}$.
\[\Pr(A_{1}|A_{2})\leq\sum_{k,m,n}\Pr(|\hat{\mu}^k(m,n)-\mu(m,n)|\geq \epsilon|A_2),\] 
where the inequality follows from union bound.
To show that $\Pr(A_{1}|A_{2})\leq \frac{\delta}{2}$, it suffices to show that $\Pr(|\hat{\mu}^k(m,n)-\mu(m,n)|\geq \epsilon|A_2)\leq \frac{\delta}{2MK(\beta+1)}$ which follows from Lemma \ref{thm:esterr} in the appendix with $\delta \gets \frac{\delta}{2MK(\beta+1)}$ for $n\geq 2$ and follows from Hoeffding's inequality for $n=1$. 
\end{IEEEproof}

\begin{lem}\label{sto:K}
For any $\delta$, if we run the estimation phase of the algorithm for $T_0 \geq  \lceil\frac{M^2\exp2(\frac{K-1}{M-1})}{2(0.49)^2}\ln(\frac{2}{\delta})\rceil$ rounds, then with probability at least $1-\delta$, we have $\hat{K}= K$. 
\end{lem}

\begin{IEEEproof}
Probability of collision for a user at any time is given by 
\[p=1-\Pr(\text{No collision})=1-\sum_{\text{channels}}\frac{1}{M}(1-\frac{1}{M})^{K-1}=1-(1-\frac{1}{M})^{K-1}.\] 

Let $\hat{p_{t}}=\frac{\sum_{\tau}1\{\text{collision at time }\tau\}}{t}$. We have $E[\hat{p_{t}}]=p$ and we can use Hoeffdings inequality
since collision at each time-slot is independent. Thus if $t\geq\frac{\ln(\frac{2}{\delta})}{2\epsilon_{2}^{2}}$,
with probability greater than $1-\delta$, we have 
$\hat{|p_{t}}-p|\leq\epsilon_{2}.$

We have $\hat{K}=\text{round}(\frac{\ln(1-\hat{p}_{t})}{\ln(1-\frac{1}{M})}+1)$ and $K=\frac{\ln(1-p)}{\ln(1-\frac{1}{M})}$. In order to show $\hat{K}=K$, it suffices to show
\[
|\hat{K}-K|=|\frac{\ln(\frac{1-\hat{p}_{t}}{1-p})}{\ln(1-\frac{1}{M})}|\leq0.49,
\]
which is equivalent to showing 
\[
(1-p)(1-(1-\frac{1}{M})^{-0.49})\leq\hat{p_{t}}-p\leq(1-p)(1-(1-\frac{1}{M})^{0.49}).
\]
It suffices to show 
\[
\epsilon_{2}\leq(1-p) \min\{|(1-(1-\frac{1}{M})^{-0.49})|,|(1-(1-\frac{1}{M})^{0.49})|\}.
\]
We have 
\[
|(1-(1-\frac{1}{M})^{-0.49})|=(1+\frac{1}{M-1})^{0.49}-1\geq\frac{0.49}{M-1})
\]
 and 
\[
(1-(1-\frac{1}{M})^{0.49})\geq\frac{0.49}{M}
\]
where the inequalities follow from the Bernoulli inequality, $(1+x)^r\leq 1+xr$ for $0\leq r\leq 1$ and $x\geq -1$.

We have from  $(1-\frac{1}{x})^{x-1}\geq \frac{1}{\exp(1)}$ for $x\geq 1$,
\[
1-p=(1-\frac{1}{M})^{K-1}\geq\frac{1}{\exp(\frac{K-1}{M-1})}.
\]

Hence, we choose $\epsilon_{2}\leq\frac{0.49}{M\exp(\frac{K-1}{M-1})}$.

\end{IEEEproof}

\subsubsection{Allocation phase}
We now find bounds on the expected regret during each fixing phase, given that the estimates of $\mu(m,f(m))$ and $K$ are accurate.
\begin{lem}\label{thm:sto_fixing}
The expected regret accumulated by the system during a fixing phase is upper bounded by
\vspace*{-0.05 in}
\[ K^{2}M\text{exp}\left(\frac{K-1}{M-1}\right).\]
\end{lem}

\begin{IEEEproof}
Let ${\cal  M}_t$ denote the set of unfixed arms at time $t$.
 Probability of user $k$ being fixed at time $t$ is given by,

$\Pr(\text{User }k\text{ being fixed})$
\begin{eqnarray*}
 & = & \sum_{m\in {\cal  M}_t}\Pr(\text{Choosing arm }m) \Pr(\text{Being fixed}|\text {arm }m)\\
 & = & \sum_{m\in {\cal  M}_t}\frac{1}{M} \Pr(\text{At most \ensuremath{f_{m}^{\ast}-1} users choose arm }m)\\
 & = & \sum_{m\in {\cal  M}_t}\frac{1}{M}\sum_{i=0}^{\ensuremath{f_{m}^{\ast}-1}}{K-1 \choose i}\left(\frac{1}{M}\right)^{i}\left(1-\frac{1}{M}\right)^{K-1-i}\\
 & \underset{(a)}{\geq} & \frac{1}{M}\left(1-\frac{1}{M}\right)^{K-1}=\frac{1}{M}\left(1-\frac{1}{M}\right)^{(K-1)*(M-1)/(M-1)}\\
 & \underset{(b)}{\geq} & \frac{1}{M}\frac{1}{\text{exp}(\frac{K-1}{M-1})}
\end{eqnarray*}
where $(a)$ follows because we only consider one term in the each
of the summations with $i=0$, and $(b)$ follows from  $(1-\frac{1}{x})^{x-1}\geq \frac{1}{\exp(1)}$ for $x\geq 1$.
Thus for any user $k$, the expected fixing time is given by 
\[
\mathbb{E}[t^{k}_{f}]=\frac{1}{p(\text{User }k\text{ being fixed})}\leq M\text{exp}\left(\frac{K-1}{M-1}\right)
\]
and thus the  regret during the fixing phase is given by,
\[
\mathbb{E}\left[\sum_{k}\sum_{t=T_0+1}^{\max_kt^k_f}R^k_{t}\right]\leq \mathbb{E}\left[K\text{max }{t^k_f}\right]\leq\mathbb{E}\left[K\sum_{k=1}^{K}t^k_f\right]\leq K^{2}T_f,
\]
where $R^k_{t}$ denotes the regret incurred by user $k$ at time $t$ and we have $R^k_{t}\leq 1$ by our assumption on the reward distribution.
\end{IEEEproof}
\vspace*{-0.08 in}
\subsection*{Analysis for $K\leq M$}
For the case where $K\leq M$, there is no need for clustering. We only need the estimates for $\mu(m,1)$, and all users individually choose the best $K$ channels. This reduces to the ``musical chairs" algorithm and the analysis can be found in \cite{RosenskiSS15mc}. After fixing on a channel during the first allocation time, after every $T_x$ time units, each user switches to the next channel among the $K$ best channels. 

\subsubsection{Main Result}
We now present the upper bound on the expected regret incurred by the users employing Algorithm \ref{sto:main}.
\begin{thm}\label{thm:main}
For any fixed $\epsilon$ and $\delta\in (0,1)$,  with probability greater than $ 1 - \delta $, the expected regret for $ K $ users using Algorithm  \ref{sto:main} with $ M $  arms for $T $ rounds, with parameter $T_0 = \left\lceil  \frac{32\exp(\frac{K-1}{M-1})M}{\epsilon^2}\ln \frac{2MK\beta(\beta+1)}{\delta} \right\rceil$, $T_c\sim O(T_0)$ and any $N_0$, is given by 
\[\mathbb{E}[R(T)] \leq K(T_0+T_c)+N_0K^{2}M\text{exp}\left(\frac{K-1}{M-1}\right),\]
i.e., $\mathbb{E}[R(T)] \sim O(1)$ in $T$.
\end{thm}

\begin{IEEEproof}
The expected regret is due to regret during the estimation phase as well as the allocation phase.

 Let $T_f$ denote the time taken for all the users to fix. 
\begin{eqnarray*}
\mathbb{E}[R(T)] & \leq &  K(T_0+T_c)+KN_0\mathbb{E}[T_f]\\
 & \leq & K(T_0+T_c)+N_0K^{2}M\text{exp}\left(\frac{K-1}{M-1}\right)
\end{eqnarray*}

From Lemmas \ref{sto:mu} and \ref{sto:K}, we have the correct estimates for $\mu(m,f(m))$ and $K$ with high probability,  with an estimation phase of $T_0+T_c$ time units. Thus, $K(T_0+T_c)$ corresponds to the regret accumulated system-wide during the estimation phase. Here $T_c$ denotes the time used for running the $\alpha$-approximation algorithm for clustering.  In the allocation phase, the regret in the system is accrued only during the $N_0$ number of fixing phases. From Lemma \ref{thm:sto_fixing}, the regret in each fixing phase is $K^{2}M\text{exp}(\frac{K-1}{M-1})$. 
\end{IEEEproof}

\subsection{Dynamic case}\label{sto:dynm}

We now extend the results to a dynamic system with a changing number of users. The key idea is to run Algorithm 1 repeatedly across epochs. However, in order to obtain a sub-linear regret bound, we need to impose  some restrictions on the number of epochs, and on the way users enter or leave the system. It is easy to see that the number of epochs $N_e$ must be sub-linear in time to have sub-linear regret in the system. We restrict the number of users entering and leaving the system until time $t$, which we denote by $\Delta_t$ to be $O(t^{\zeta})$ where $\zeta<\frac{1}{2}$. We note that this is different from \cite{RosenskiSS15mc} where the time horizon is fixed and known, and there is also a restriction on when users can enter or leave the system. In our model, the dynamic scenario also includes the case where $K_t$ can go from greater than $M$ to less than $M$, and vice-versa.

Let $K_t$ denote the number of active users at time $t$, where $\frac{K_t}{M}\leq \beta$. Note that all the theorems in subsection \ref{sto:analysis} follow for the dynamic case with $K_t\leq M\beta$.
We choose the starting epoch length $\tau$ to be greater than or equal to $ T_0^{(1)}+T_c^{(1)}+N_0(T_x+T_f)$. We run Algorithm 1 for time $\tau$, $2 \tau$, $3 \tau$,  and so on. The resulting algorithm is given below.


\begin{algorithm}[htb]
  \caption{Dynamic Allocation}
  \label{sto:dyn}
  \begin{algorithmic}[1]
 \For{$\tau\frac{r(r+1)}{2}\leq \tau \leq \tau\frac{(r+1)(r+2)}{2}$}
 \State Run Algorithm 1 with  $\delta^{(r)} \gets \frac{\delta}{2^{r+1}}$.
 \EndFor
  \end{algorithmic}
\end{algorithm}
\begin{thm}
With a probability greater than $1-\delta$, the expected system-wide regret after running the Algorithm \ref{sto:dyn} for $T$ rounds  where $\tau\frac{r(r+1)}{2}\leq T \leq \tau\frac{(r+1)(r+2)}{2}$ is 
\[\mathbb{E}[R(T)]\leq  M\beta[N_{e}(T_{0}^{(r)}+T_{c}^{(r)}+M\beta T_{f})+\Delta_T  {T}^{\frac{1}{2}}],\]
i.e., $\mathbb{E}[R(T)] \sim O(\Delta_T{T}^{\frac{1}{2}})$.
\end{thm}

\begin{IEEEproof}
We have $\tau\frac{r(r+1)}{2}\leq T$ which gives us $r\leq (\frac{2T}{\tau})^{\frac{1}{2}}$.
The epoch length is changing with time. The total number of epochs $N_e$ until time $T$ is $N_e\leq(r+1)\sim O({T}^{\frac{1}{2}})$.

We consider Theorem \ref{sto:main} with $\delta$ set as in the algorithm Dynamic allocation. Thus we have $T_{0}^{(r)}\sim O(\ln r)\sim O(\ln T)$.
Using union bound we show that with a probability at least $1-\delta$, we have good estimates for $\mu$ and $K_t$ over all epochs. 
\begin{eqnarray*}
\Pr( \exists \text { epoch with wrong estimate}) & \leq \\
 \vspace*{-0.15 in}
 \sum_{\text{epochs}}\Pr(\text{wrong estimate})
 & \leq & \delta\sum_{i=1}^{r+1}\frac{1}{2^{i}}\leq\delta.
 \vspace*{-0.10 in}
\end{eqnarray*}
In epochs with fixed or static users, the accumulated regret follows from Theorem \ref{sto:main}, and in epochs with dynamic users, the system incurs regret during the entire epoch.
\vspace{-0.10 in} 
\begin{eqnarray*}
\mathbb{E}[R(T)] & \leq & N_{e}(\text{Static case regret})+K_t\sum^{\Delta_T}\text{Epoch length}\\
 & \leq &  M\beta[N_{e}(T_{0}^{(r)}+T_{c}^{(r)}+M\beta T_{f})]+\Delta_T r \tau]\\
 & \leq &  M\beta[N_{e}(T_{0}^{(r)}+T_{c}^{(r)}+M\beta T_{f})+\Delta_T{T}^{\frac{1}{2}}].
\end{eqnarray*}
Thus, if  $\Delta_T$ is $O(T^{\zeta})$ where $\zeta<\frac{1}{2}$, we have sub-linear regret.
\end{IEEEproof}

\subsection{Experiments}\label{sec:exp}
In this section, our goal is to validate the performance of the estimation phase in the algorithm and show that the performance in the allocation phase does not suffer due to use of the estimated values i.e., the regret does not grow with time in the allocation phase. 

We consider a system with $K=10$ users and $M=6$ channels and the non dynamic case.  We set $T_0=1000$, $T_x = 1000$ time units and $N_0=5$ and repeat the experiment 100 times and consider the average accumulated regret. The value of $\beta$ is set to 3,  and the reward distributions are chosen to be uniform with a variance of 0.01, and means between $0$ and $1$ given below,
\[\mu=
\begin{bmatrix}
   1 &0.49 & 0.1 & 0.005     \\
    0.98       & 0.42 &0.13 & 0.002 \\
     0.97&0.5&0.12&0.009\\
    1&0.48&0.009&0.008\\
    0.92&0.43&0.1&0.001\\
    0.9&0.44&0.1&0.001
\end{bmatrix}.\]
 We compare the performance of Algorithm \ref{sto:main} with the estimated values of $\mu$ and $K$ with Algorithm \ref{sto:main}  with the true parameter values. We also show how the estimates change with number of iterations in the estimation phase $T_0$. We used the in-built MATLAB {\tt kmeans} function for clustering.

\begin{figure}[htb]
  \centering
\includegraphics[scale = 0.34]{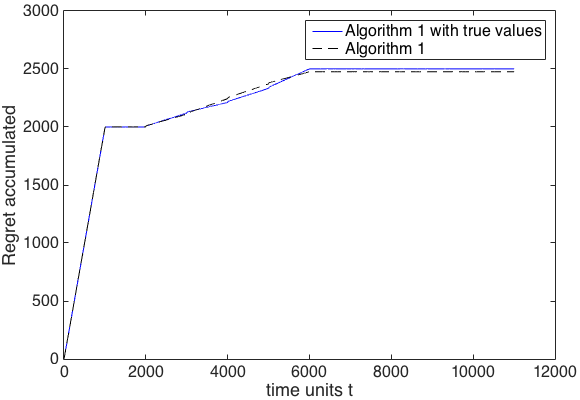}
  \caption{Accumulated regret as a function of time.}
 \label{fig:sto_regret}
 \setlength{\belowcaptionskip}{-10pt}
\end{figure}

\begin{figure}[htb]
  \centering
\includegraphics[scale = 0.4]{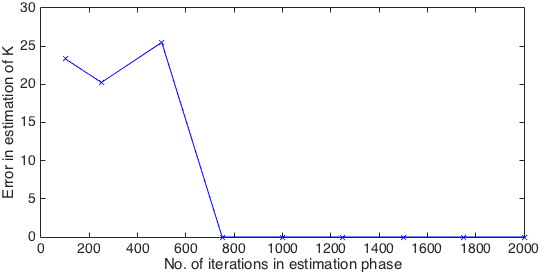}
  \caption{Error in the estimation of number of users $K$.}
 \label{fig:kerr}
  \setlength{\belowcaptionskip}{-10pt}
\end{figure}

\begin{figure}[htb]
  \centering
\includegraphics[scale = 0.4]{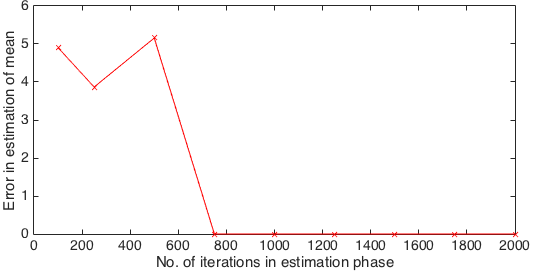}
  \caption{Error in the estimation of the mean. }
 \label{fig:muerr}
  \setlength{\belowcaptionskip}{-20pt}
\end{figure}

From Fig. \ref{fig:sto_regret}, we see that the accumulated regret grows with time during the estimation phase and remains constant during the allocation phase. Also, there is no noticeable difference between Algorithm \ref{sto:main}  with the true parameter values and the one with the estimated values. This follows because the estimates of $K$ and the mean converge to the true values within a  few iterations as shown in Fig. \ref{fig:kerr} and  Fig. \ref{fig:muerr}.

\section{Adversarial setting}

In this section, we consider the adversarial multi-user MAB model with user-dependent rewards on each channel. We present an algorithm that leads to sub-linear regret, and  extend it to the dynamic case.

\subsection{Single user MAB}\label{sec:single}

We consider the Exp3.P algorithm described in \cite{bubeck12survey} for a single user MAB in an adversarial setting.   We modify the algorithm so that the user chooses an arm and updates the probability vector only in a few time units. This modification is useful in the multi-user case, where the users may not choose an arm in each time unit due to possible collisions.
We now present a modified version of the Exp3.P algorithm, in which a new arm is chosen and the probability is updated at time units $t_1,t_2,\ldots,t_n$ such that $n\leq T$ and $\alpha = \max_{j\in [n-1]} t_{j+1}-t_j$. For each $j\in [n]$, we consider the reward over the time-period $t_{j+1}-t_j$, with the reward being normalized to lie between $0$ and $1$. 

\begin{algorithm}[h]
  \caption{Modified Exp3.P}
  \label{ad:exp3}
  \begin{algorithmic}[1]
    \State $\phi = \sqrt{\frac{\ln M}{Mn}}$,\quad$\eta =0.95 \sqrt{\frac{\ln M}{Mn}}$ and $\gamma = 1.05\sqrt{\frac{M\ln M}{n}}$.
  \State Initial probability distribution $p_0=(\frac{1}{M},\ldots,\frac{1}{M})$.
 \For {$j=1,\ldots,n$}
 \State $a_{j}\sim p_j$, remain on arm for next $t_{j+1}-t_{j}$ time units
\State Compute reward as $g'_{j}(i)=\frac{\sum_{t_{j}\leq t\leq t_{j+1}} {g}_{t}(i)}{t_{j+1}-t_{j}}$
and the estimated gain for each arm as 
\[\tilde{g}_{j}(i)=\frac{g'_{j}(i)\mathbbm{1}_{a_j=i}+\phi}{p_{j}(i)}\] 
and update the cumulative gain $\tilde{G}_{j}(i)=\sum_{s=1}^{j}\tilde{g}_{s}(i)$
\State Calculate 
     $p_{j+1} = (p_{j+1}(1),...,p_{j+1}(M))$ where\[p_{j+1}(i)=(1-\gamma)\frac{\exp(\eta\tilde{G}_{j}(i))}{\sum_{m=1}^{M}\exp(\eta\tilde{G}_{j}(m))}+\frac{\gamma}{M}\]
\EndFor
  \end{algorithmic}
\end{algorithm}

\begin{thm}\label{thm:gen_ad}
The expected regret of Modified Exp3.P algorithm (Algorithm \ref{ad:exp3}) until time $T$ is given by \\
$\mathbb{E}\left[\sum_{t=1}^{T}g_{t}(m) - g_{t}(a_t)\right]$\vspace{-0.1in}
\begin{equation}\label{eqn:ad}
\leq \underset{m\in[M]}{\max} \mathbb{E}\left[\sum_{t=1}^{T}(g_{t}(m) -  g_{t}(a_t))\right]\leq\alpha\sqrt{n}h(M)
\end{equation}
where $h(M)=5.15\sqrt{M\ln M}+\sqrt{\frac{M}{\ln M}}$, and does not depend on $T$ and $n\leq T$.
\end{thm}


\begin{IEEEproof}
We have
\begin{equation}\label{eqn:adproof}
\mathbb{E}\left[\sum_{t=1}^{T}(g_{t}(m) - g_{t}(a_t))\right]\leq \alpha \mathbb{E}\left[\sum_{j=1}^{n}(g'_{j}(m) - g'_{j}(a_j))\right],\end{equation}
where $g'_{j}(m) = \frac{\sum_{t_{j}\leq t\leq t_{j+1}} {g}_{t}(m)}{t_{j+1}-t_{j}}$.
Using \eqref{eqn:adproof}, and noting that until time $T$ we consider $n$ time units, the proof follows from the regret bound for Exp3.P given in \cite{bubeck12survey}. 
\end{IEEEproof}

\subsection{Multi-user MAB: Algorithm}\label{ad:alg}

We now consider the multi-user adversarial bandits under a known finite horizon $T$, and propose an algorithm which when employed by all users independently leads to sub-linear regret.

In a multi-user adversarial system, every time $t$ that a user $k$ chooses an arm according to a certain probability distribution $p^k_t$ to randomize against the adversary, there is a possibility for collision with other users. Hence there is a need for a collision resolution mechanism, so that the regret does not grow linearly with time. Instead of choosing an arm every time unit, a user chooses an arm only a sub-linear number of times until $T$ ( e.g., $T^y$ where $y<1$).
The goal is to randomize sufficient number of times so as to counteract the adversary, while making sure that the regret due to collisions does not become large.

We propose an algorithm (Algorithm \ref{ad:main}) that combines the modified Exp3.P algorithm (Algorithm  \ref{ad:exp3}) with a collision resolution mechanism with  $y<1$.  In the analysis in Section \ref{ad:analysis}, we pick $y=\frac{1}{2}$ which is large enough to maintain the sub-linear regret achieved by the modified Exp3.P algorithm but small enough so that the regret due to collisions is sub-linear as well. 

In every time-interval of length $T^{1-y}$, we first have a collision resolution phase. Each user chooses a channel with probability $p^k_t$. A user settles or ÔfixesÕ on a channel if at any time the user finds a channel without collision. Once a user settles on a channel, the user keeps transmitting on the channel until the end of the time-interval of length $T^{1-y}$. The system incurs regret until all $K$ users have settled on $K$ channels, and we call this duration the \emph{fixing time}. The remaining part of the algorithm corresponds to each of the $K$ users employing the modified Exp3.P algorithm, where they choose a channel once every $T^y$ time units.

\begin{algorithm}[h]
  \caption{}
  \label{ad:main}
  \begin{algorithmic}[1]
    \State $\phi = \sqrt{\frac{\ln M}{MT^y}}$,\quad$\eta =0.95 \sqrt{\frac{\ln M}{MT^y}}$ and $\gamma = 1.05\sqrt{\frac{M\ln M}{T^y}}$.
  \State The initial probability distribution $p^k_0=(\frac{1}{M},\ldots,\frac{1}{M})$
 \For {$t=\text{ multiples of } \frac{T}{T^y}$}
 \For {$t'=1$ to $T^{1-y}$}
 \State $a^k_{t'}\sim p^k_t$
 \If {no collision}
 \State break 
   \EndIf
  \EndFor
 \State Choose action $a^k_{t'}$ for next $T^{1-y}-t'$  time units
\State Compute reward as $g'^k_{t}(i)=\frac{\sum {g}^k_{t}(i)}{T^{1-y}-t'}$
and the estimated gain for each arm as 
\[\tilde{g}^k_{t}(i)=\frac{g'^k_{t}(i)\mathbbm{1}_{a^k_{t'}=i}+\phi}{p^k_{t}(i)}\] 
and update the cumulative gain $\tilde{G}^k_{t}(i)=\sum_{s=1}^{t}\tilde{g}^k_{s}(i)$
\State Calculate 
     $p^k_{t+1} = (p^k_{t+1}(1),...,p^k_{t+1}(M))$ where
     \begin{equation}\label{eqn:p}p^k_{t+1}(i)=(1-\gamma)\frac{\exp(\eta\tilde{G}^k_{t}(i))}{\sum_{m=1}^{M}\exp(\eta\tilde{G}^k_{t}(m))}+\frac{\gamma}{M}\end{equation}
\EndFor
  \end{algorithmic}
\end{algorithm}

\subsection{Multi-user MAB: Analysis}\label{ad:analysis}
In this subsection, we first consider the regret due to the collision resolution phase, then the regret due to the modified Exp3.P part of Algorithm \ref{ad:main}, and then combine them to find an upper bound on the system-wide regret incurred when each user independently employs Algorithm \ref{ad:main}.

\subsubsection{Regret during collision resolution}
\begin{thm}\label{thm:fixing}
The expected regret accumulated by the system during a collision resolution phase  is upper bounded by
 \[\frac{K^{2}M^{K}}{\gamma}\leq\frac{K^{2}M^{K}T^{\frac{y}{2}}}{\sqrt{M\ln M}}.\]
\end{thm}
\begin{IEEEproof}
We first note from equation \eqref{eqn:p} that the probability of choosing any channel by any user is at least $\frac{\gamma}{M}$. Let $\rho^k_t = \max_{m}p_{t}^k(m)$, which implies that $\rho^k_t \geq \frac{1}{M}$. Let ``maximal" refer to the channel that has the highest probability of being chosen by that particular user. 
Thus, each user can be associated with one channel such that probability of choosing it is greater than $ \frac{1}{M}$. Since $K\leq M$, for each user, there exists at least one channel such that it not the maximal channel for any of the remaining $K-1$ users.
Note that even when some users fix or settle on a channel, and there are both unfixed channels and unfixed users in the system, we can still find an unfixed channel such that it is not the maximal channel for the remaining unfixed users. 

Based on the above discussion, we define the event
$B_k$ to be the event where all unfixed users except user $k$ choose their maximal arm, and user $k$ chooses an unfixed arm that is not the maximal arm for any other unfixed users. 

Let ${\cal M}_{t}$ denote the set of unfixed arms at time $t$. The probability of any user $k$ being fixed at time $t$ is given by,

$\Pr\{\text{User }k\text{ being fixed}\}$
\vspace{-0.03 in}
\begin{eqnarray*}
 & = & \sum_{m\in {\cal M}_{t}} \Pr\{\text{User $k$ is the only unfixed user on arm }m\}\\
  & \geq &\Pr(B_k)\\
   & \geq  &(\Pi_{i\in [K],i\neq k}\rho^i_t) \underset{m\in {\cal M}{t}}{\min}p^k_{t}(m)\\
  & \geq & \frac{\gamma}{M}(\frac{1}{M})^{K-1}= \frac{\gamma}{M^K}.
\end{eqnarray*}

The remainder of the proof follows in a similar manner as the proof of Theorem \ref{thm:sto_fixing}.
\end{IEEEproof}

\subsubsection{Regret due to Modified Exp3.P}
We now bound the regret incurred by the users using Algorithm \ref{ad:main} during the time the users are not in the collision resolution phase. This corresponds to each of the $K$ users independently employing the modified Exp3.P algorithm introduced in subsection \ref{sec:single}.


In Algorithm \ref{ad:main}, when the users are not in the collision resolution phase, each user employs modified Exp3.P with $n=T^{y}$ and $\alpha =T^{1-y}$. Using the result of Theorem \ref{thm:gen_ad} for $K$ users, for any distinct set ${\cal K} \subseteq [M]$ consisting of $K$ arms, 
\[\mathbb{E}\left[\sum_{t\notin \text{coll. phase}}\left(\sum_{i\in {\cal K}}g^{k}_{t}(i) - \sum_{k=1}^{K} g^{k}_{t}(a^k_t)\right)\right]\leq KT^{1-\frac{y}{2}}h(M).\]
\vspace{-0.12in}
Thus, 
 \begin{equation}\label{eqn:exp}\max_{\cal K}\mathbb{E}\left[\sum_{t\notin \text{coll. phase}}\left(\sum_{i\in {\cal K}}g^{k}_{t}(i) - \sum_{k=1}^{K} g^{k}_{t}(a^k_t)\right)\right]\leq h(M)KT^{1-\frac{y}{2}}\end{equation}
where $h(M)=5.15\sqrt{M\ln M}+\sqrt{\frac{M}{\ln M}}$, and does not depend on $T$.

\subsubsection{Main Result}

We now present the upper bound on the expected regret incurred by the users employing Algorithm \ref{ad:main}.

\begin{thm}\label{thm:main}
The expected regret of $ K $ users using Algorithm \ref{ad:main} with $M$ arms for $T $ time units,  is given by 
\[\mathbb{E}[R(T)] \leq T^{\frac{3}{4}}h'(M,K)\]
 where $h'(M,K)= K\left(5.15\sqrt{M\ln M}+\sqrt{\frac{M}{\ln M}}+\frac{KM^{K}}{\sqrt{M\ln M}}\right)$, and does not depend on $T$.
Thus, $\mathbb{E}[R(T)]\sim O(T^{\frac{3}{4}})$.
\end{thm}

\begin{IEEEproof}
The expected regret is due to collision resolution phase as well as the modified Exp3.P algorithm which is played a sub-linear number of times. Let $T_f$ denote the time taken for all the users to fix.
\begin{eqnarray*}
\mathbb{E}[R(T)] & \leq & T^{y} K\mathbb{E}[T_f]+(T^{1-y}-\mathbb{E}[T_f])h(M)T^{\frac{y}{2}}\\
 & \leq & \frac{K^{2}M^{K}}{\sqrt{M\ln M}}T^{\frac{3y}{2}}+ KT^{1-\frac{y}{2}}h(M)\\
 & \sim & O(T^{{\frac{3y}{2}}}+T^{1-\frac{y}{2}})
\end{eqnarray*}
where the inequalities follow from Theorem \ref{thm:fixing} and equation \eqref{eqn:exp}, and $h(M)=5.15\sqrt{M\ln M}+\sqrt{\frac{M}{\ln M}}$.
If we choose $y$ such that ${\frac{3y}{2}}=1-\frac{y}{2}$, we have $y=\frac{1}{2}$ which gives us
\[
\mathbb{E}[R(T)]\leq  T^{\frac{3}{4}}K\left(\frac{KM^{K}}{\sqrt{M\ln M}}+h(M)\right).
\]
\end{IEEEproof}

\subsection{Unknown time horizon}
In this subsection, we extend the results to the case of unknown time horizon. Each user considers some known time $\tau$ greater than the expected fixing time for the system and runs Algorithm \ref{ad:main}. Once the user reaches the end of time $\tau$, the user continues to use  Algorithm \ref{ad:main} with a time-period of length $2\tau$. In this way when the user reaches the end of the previous time-period, the user doubles it and continues with Algorithm \ref{ad:main}. Let $T$ be such that $\tau+2\tau+\ldots+2^r\tau \leq T\leq \tau+2\tau+\ldots+2^{(r+1)}\tau$, equivalently $2^{(r+1)}\tau\leq T+\tau< 2^{(r+2)}\tau$.

\begin{algorithm}[htb]
  \caption{}
  \label{ad:dyn}
  \begin{algorithmic}[1]
 \For{$(2^{(r+1)}-1)\tau\leq T< (2^{(r+2)}-1)\tau$}
 \State Run Algorithm 1 with time-period $2^{r+1}\tau$
\EndFor
  \end{algorithmic}
\end{algorithm}

\begin{thm}\label{thm:inf}
The expected regret from using Algorithm \ref{ad:dyn} for $T$ time units where $(2^{(r+1)}-1)\tau\leq T< (2^{(r+2)}-1)\tau$ is 
\[\mathbb{E}[R(T)]\leq h'(M,K)\frac{(2(T+\tau))^{\frac{3}{4}}}{2^{\frac{3}{4}}-1}\]
where $h'(M,K)= K\left(5.15\sqrt{M\ln M}+\sqrt{\frac{M}{\ln M}}+\frac{KM^{K}}{\sqrt{M\ln M}}\right)$ and does not depend on $T$.
Thus, $\mathbb{E}[R(T)] \sim O( {T}^{\frac{3}{4}})$.
\end{thm}
\begin{IEEEproof}
We have $2^{(r+1)}\tau\leq T+\tau$.
Using Theorem \ref{thm:main}, the regret up to  time $T$ bounded as follows:
\begin{eqnarray*}
\mathbb{E}[R(T)] & \leq & h'(M,K)(\tau^{\frac{3}{4}}+(2\tau)^{\frac{3}{4}}+\ldots +(2^{r+1}\tau)^{\frac{3}{4}})\\
 & = & h'(M,K) \tau^{\frac{3}{4}}\frac{(2^{(r+2)\frac{3}{4}}-1)}{2^{\frac{3}{4}}-1}\\
  & \leq & h'(M,K)  \frac{(2(T+\tau))^{\frac{3}{4}}-\tau^{\frac{3}{4}}}{2^{\frac{3}{4}}-1}.
\end{eqnarray*}
\end{IEEEproof}

Note that each user only needs knowledge of $K$ in order to fix on an initial $\tau$ such that $\tau\geq \mathbb{E}{T_f}$, where ${T_f}$ is the fixing time for all the users in the system. Furthermore, $\tau$ can be chosen even  without the knowledge of $K$ by simply replacing $K$ by $M$, and the analysis follows because $K\leq M$. 
\vspace{-0.12 in}
\subsection{Dynamic case}\label{ad:dynm}
In this subsection, we extend the results to a dynamic system with a changing number of users. Consider a system  which starts with $K$ users, and in which users leave the system once they are done with their transmission. It is easy to see that  Algorithm \ref{ad:dyn} in this case leads to system-wide regret of the order $O(T^\frac{3}{4})$ over a time horizon $T$. 

Let us now consider a dynamic system where users enter and leave the system over time.
In order to use Algorithm \ref{ad:dyn} to obtain a sub-linear regret bound, we need to impose some restrictions on the number of users that have entered the system until time $t$, which we denote by $\kappa_t$. It is easy to see that the number of epochs in which users enter the system must be sub-linear in time to have sub-linear regret in the system. We restrict the number of users entering the system $\kappa_t$ to be $O(t^{\zeta})$ where $\zeta<\frac{1}{2}$. 
We note that this is similar to the dynamic case in \cite{bande18} where there is a restriction on the number of users entering and leaving the system. 

Let $K_t$ denote the number of active users at time $t$. Note that even in the dynamic scenario, we still retain the assumption of having $K_t\leq M$ in the system.

\begin{thm}
The expected system-wide regret from using Algorithm \ref{ad:dyn} for $T$ time units where $(2^{(r+1)}-1)\tau\leq T< (2^{(r+2)}-1)\tau$  with the number of users entering the system $\kappa_T\sim O(T^{\zeta})$, with $\zeta<\frac{1}{2}$, is given by
\[\mathbb{E}[R(T)]\leq h'(M,M) \frac{(2(\tau+T))^{\frac{3}{4}}}{2^{\frac{3}{4}}-1}+M\kappa_T  {T}^{\frac{1}{2}}\]
where $h'(M,M)= M\left(5.15\sqrt{M\ln M}+\sqrt{\frac{M}{\ln M}}+\frac{M^{M+1}}{\sqrt{M\ln M}}\right)$ and does not depend on $T$.
Thus, $\mathbb{E}[R(T)] \sim O( T^{\frac{3}{4}}+\kappa_T T^\frac{1}{2})$.
\end{thm}

\begin{IEEEproof}
We have $2^{(r+1)}\tau\leq \tau+T$. 
In epochs where no users enter the system, the regret can be bound by Theorem \ref{thm:inf}, and in epochs with new users, the regret accumulates through the entire epoch. The epoch length is upper bounded by $(2^{(r+1)}\tau)^\frac{1}{2}$, since $y=\frac{1}{2}$ from Theorem \ref{thm:main}.
The regret up to time $T$ bounded as follows:
\begin{eqnarray*}
\mathbb{E}[R(T)] & \leq & \text{Static case regret}+K_t\sum^{\kappa_T}\text{Epoch length}\\
  & \leq &  h'(M,M) \frac{(2(\tau+T))^{\frac{3}{4}}}{2^{\frac{3}{4}}-1}+M \kappa_T (2^{r+1}\tau)^{\frac{1}{2}}\\
    & \leq & h'(M,M)  \frac{(2(\tau+T))^{\frac{3}{4}}}{2^{\frac{3}{4}}-1}+M \kappa_T (\tau+T)^{\frac{1}{2}}.
\end{eqnarray*}
Thus, if  $\kappa_T$ is $O(T^{\zeta})$, with $\zeta<\frac{1}{2}$, we have sub-linear regret.
\end{IEEEproof}
\vspace{-0.15 in}
\subsection{Experiments}\label{ad:exp}
In this section, we illustrate the performance of our algorithm in a simple adversarial setting. We consider a non-oblivious adversary, i.e., an adversary whose rewards do not depend on the users' reward history.

We consider a system with known time-horizon $T$, fixed number of users $K=4$ users and $M=7$ channels.  We set $T=160000$, which gives us $T^\frac{1}{2} = 400$ time units, $\phi= 0.026$, $\eta= 0.025$ and $\gamma =0.194$ in Algorithm \ref{ad:main}. The reward distributions for the channels are drawn i.i.d from the uniform distribution $[a,1]$ where $a$ for each channel at each time unit is drawn i.i.d  from the uniform distribution $[0.2,1]$.

\begin{figure}[htb]
  \centering
\includegraphics[scale = 0.45]{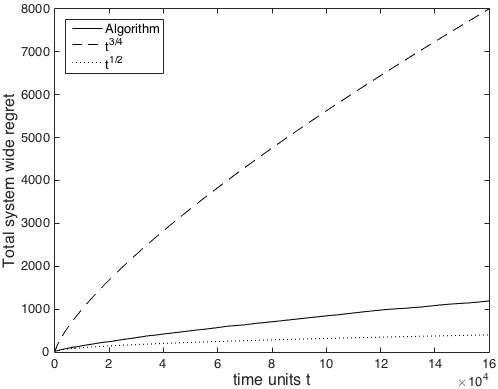}
  \caption{Accumulated regret as a function of time.}
 \label{fig:ad_regret}
 \setlength{\belowcaptionskip}{-10pt}
\end{figure}

We repeat the experiment 100 times and consider the average accumulated regret with time. From Fig. \ref{fig:ad_regret}, we see that the regret grows with time at a rate much lower than $T^\frac{3}{4}$, but higher than $T^\frac{1}{2}$, the expected regret in the single user case. 

\begin{remk}
We note that Algorithm \ref{ad:main} can be used for a stochastic multi-user MAB with user-dependent rewards to achieve a sub-linear regret of order $O(T^\frac{3}{4})$. While the regret is much higher than in \cite{kalathil2014decentralized}, our algorithm does not rely on communication between the users and can also deal with a dynamic number of users in the system.
\end{remk}

\begin{remk}
In the adversarial case, there is randomization in the selection of a channel, with $\tau^{\frac{1}{2}}$ being equivalent to $T_x$, and hence each user does not transmit on a channel for a very long time. Thus, fairness is achieved without enforcing a strict duration $T_x$ for each transmission.
\end{remk}

\section{Conclusions}\label{sec:conc}
We modeled the dynamic spectrum allocation problem as a multi-user MAB with no communication among the users. We first considered a stochastic MAB model with rewards on the channel being the same for all users, and then an adversarial MAB model with user-dependent rewards. We showed that the proposed algorithms in both scenarios achieve sub-linear regret. We provided simulation results to show that the algorithms perform well in practice when the number of users is fixed. We also extended our algorithms to the dynamic case and showed that the algorithms continue to achieve sub-linear regret. 
It is of interest to develop algorithms in other variants of the multi-user MAB setting. For example, a system with user-dependent rewards, under the stochastic as well as the adversarial settings, without any user communication, when there are more users than channels in the system.

\bibliographystyle{IEEEtran}
\bibliography{MeghanaRef}

\section{Appendix}

We present a lemma that ensures a certain number of observations from each distribution during the estimation phase of length $T_0$.

\begin{lem}\label{thm:num}
If $T_0 = \left\lceil  \frac{32\exp(\frac{K-1}{M-1})M}{\epsilon^2}\ln \frac{2MK\beta(\beta+1)}{\delta} \right\rceil$, then all users using Algorithm \ref{sto:main} have at least $\frac{16}{\epsilon^2}\ln \frac{2MK\beta(\beta+1)}{\delta}$ observations of each reward distribution on each arm with probability greater than $ 1 - \frac{\delta}{2} $.
\end{lem}

\begin{IEEEproof}
Let $ A_{k,m,n} (t) = I \left\{ \mbox{player $ k $ observed arm $ m $ with $n$ users at round $ t $} \right\} $. Note that for any round $t $ and any $ k,m,n $ we have that 
\[ \Pr \left(A_{k,m,n} (t)=1 \right) = \frac{1}{M}{K-1\choose n-1}  \left(1 - \frac{1}{M} \right)^{K-n} \left(\frac{1}{M} \right)^{n-1}\] $ \implies \mathbb{E} \left[ A_{k,m,n} (t) \right] = \frac{1}{M}{K-1\choose n-1}  \left(1 - \frac{1}{M} \right)^{K-n} \left(\frac{1}{M} \right)^{n-1}\geq \frac{1}{M}   \left( 1 - \frac{1}{M} \right)^{K-1} \geq \frac{1}{M\exp(\frac{K-1}{M-1})} $ for all $M > 1$.

where the last inequality follows from  $(1-\frac{1}{x})^{x-1}\geq \frac{1}{\exp(1)}$ for $x\geq 1$.

We have,
\begin{eqnarray*}
\Pr\left(\exists k,m,n\text{ s.t. }\sum_{t=1}^{T_{0}}A_{k,m,n}(t)\leq\frac{1}{2}T_{0}\mathbb{E}[A_{k,m,n}(t)]\right) & \leq & \sum_{k}\sum_{m}\sum_{n}\Pr\left(\sum_{t=1}^{T_{0}}A_{k,m,n}(t)\mbox{\ensuremath{\leq\frac{1}{2}  T_{0} \mathbb{E}[A_{k,m,n}(t)\mbox{]}}}\right)\\
 & \leq & \sum_{k}\sum_{m}\sum_{n}\exp\left(\frac{-\frac{1}{4}T_{0}\mathbb{E}[A_{k,m,n}(t)]}{2}\right)\\
 & = & K(\beta+1) M\exp\left(\frac{-\frac{1}{4}T_{0}\mathbb{E}[A_{k,m,n}(t)]}{2}\right)
\end{eqnarray*}
where the first inequality follows from union bound and the second inequality follows from Chernoff bound. Note that for a particular $k,m$ and $n$, $ A_{k,m,n} $ is i.i.d across $t$, since all users are choosing channels uniformly at random.

In order for this probability to be upper bounded by $ \frac{\delta}{2} $ we need:
\begin{gather*}
K(\beta+1)   M   \exp\left(\frac{-\frac{1}{4}   T_0   \mathbb{E} \left[ A_{k,m,n} (t) \right] }{2} \right) < \frac{\delta}{2}\\
\implies T_0 > \frac{1}{8   \mathbb{E} \left[ A_{k,m,n} (t) \right]}   \ln \left( \frac{2K(\beta+1)   M}{\delta} \right).
\end{gather*}
We have shown that if $ T_0 > \frac{1}{8   \mathbb{E} \left[ A_{k,m,n} (t) \right]}  \ln \left( \frac{2K(\beta+1) M}{\delta} \right) $ then w.p. $ \geq 1 - \frac{\delta}{2}$ we have $ \forall k,m,n $ the number of observations player $ k $ has of arm $ m $ with $n$ users, $ \sum_{t=1}^{T_0} A_{k,m,n} \left( t \right) > \frac{1}{2}   T_0   \mathbb{E} \left[ A_{k,m,n} (t) \right] $.\\

We also need the total number of observations each player has of each arm to be at least $\frac{16}{\epsilon^2}\ln \frac{2MK\beta(\beta+1)}{\delta}$,
i.e.
\begin{gather*}
\sum_{t=1}^{T_0} A_{k,m,n} \left( t \right) > \frac{1}{2}   T_0   \mathbb{E} \left[ A_{k,m,n} (t) \right] \geq  \frac{16}{\epsilon^2}\ln \frac{2MK\beta(\beta+1)}{\delta} \\
\implies T_0 \geq    \frac{2}{\mathbb{E} \left[ A_{k,m,n} (t) \right]}  \frac{16}{\epsilon^2}\ln \frac{2MK\beta(\beta+1)}{\delta}.\end{gather*}
So we have two constraints on $ T_0 $, which gives us:\\
\[T_0 = \left\lceil \max \left\{ \frac{1}{8   \mathbb{E} \left[ A_{k,m} (t) \right]}   \ln \left( \frac{ 2K(\beta+1) M}{\delta} \right), 2   \frac{1}{\mathbb{E} \left[ A_{k,m} (t) \right]}  \frac{16}{\epsilon^2}\ln \frac{2MK\beta(\beta+1)}{\delta}  \right\} \right\rceil\] 
which can be further simplified to 
\[T_0 = \left\lceil  \frac{32\exp(\frac{K-1}{M-1})M}{\epsilon^2}\ln \frac{2MK\beta(\beta+1)}{\delta} \right\rceil\] 

\end{IEEEproof}
\subsection{Clustering}

Let $N$ points $\{x_i,\ldots,x_N\}$ be drawn independently from $\beta$ distributions with mean $\mu_r$ where $r\in [\beta]$. Let number of samples drawn from  distribution with mean $\mu_r$ be denoted by $n_r$ and the separability condition \eqref{eqn:sep} is satisfied. Additional notation used is introduced in Table \ref{tab:not2}.

We now present an additional separability condition which is useful in order to prove some clustering results.
For any $m\in[M]$ and $r,s\in [\beta]$,
\begin{equation}\label{eqn:sep_clus}|\mu(m,r)-\mu(m,s)|\geq c\phi_{\ast}(\frac{1}{n_s}+\frac{1}{n_r}),\end{equation}
where
$\phi_{\ast}=\sum_i |x_i-\mathbb{E}(x_i)|$ and $c$ is a constant.

\begin{table}[htb]
\begin{center}
\begin{tabular}{ |c |c |} 
\hline
$\Delta_s$ & $|\mu_s-\nu_s|$ \\ 
\hline
$\gamma$ & $\max_{s,r\neq s}\frac{\Delta_s}{|\mu_r-\mu_s|}$  \\ 
\hline
$\{{\cal T}_s\}_{s\in[\beta]}$ & True partition of the samples $X $ \\ 
\hline
$n_s$ & $|{\cal T}_s|$\\ 
\hline
$\phi_{\ast}$ &$\sum_i |x_i-\mathbb{E}(x_i)|$\\ 
\hline
$g(S)$ &$\frac{1}{|S|}\sum_{i\in S} x_i$\\ 
\hline
$\rho_{in}^{s}$ & Fraction of points misclassified as cluster $s$ $\frac{\sum_{r\neq s} |{\cal T}_r\cap S_s|}{n_s}$ \\ 
\hline
$\rho_{out}^{s}$ & Fraction of misclassified points in cluster $s$ $\frac{\sum_{r\neq s} |{\cal T}_s\cap S_r|}{n_s}$ \\ 
\hline
\end{tabular}
\end{center}
\caption{Notation.}
\label{tab:not2}
\end{table}

We first present the following lemma which describes the relationship between the separability conditions \eqref{eqn:sep} and \eqref{eqn:sep_clus}.


\begin{lem}
If the separability condition \eqref{eqn:sep} is satisfied and $N=T_0= \frac{32\exp(\frac{K-1}{M-1})M}{\epsilon^2}\ln \frac{2MK\beta(\beta+1)}{\delta}$, then for any $r,s$, with high probability
\begin{equation}|\mu_r-\mu_s|\geq c\phi_{\ast}(\frac{1}{n_s}+\frac{1}{n_r}),\end{equation}
where
$\phi_{\ast}=\sum_i |x_i-\mathbb{E}(x_i)|$ and $c$ is a constant.


\end{lem}

\begin{IEEEproof}

If suffices to show that with high probability, 
\[
4M\exp(\frac{K-1}{M-1})\sqrt{\sigma^{2}+\epsilon_{2}}\geq (\frac{1}{n_r}+\frac{1}{n_s})\sum_{i\in[N]}|x_{i}-E[x_{i}]|. 
\]

From Hoeffdings, we have 
\[
\Pr(\frac{1}{N}\sum_{i\in[N]}(x_{i}-E[x_{i}])^{2}-\sigma^{2}\geq\epsilon_{2})\leq\exp(-2N\epsilon_{2}^{2})
\]

i.e., with probability greater than $1-\frac{\delta}{2MK\beta(\beta+1)}$, we have $\sum_{i\in[N]}(x_{i}-E[x_{i}])^{2}\leq N(\sigma^{2}+\epsilon_{2})$.

We have $||x||_1\leq \sqrt{N}||x||_2$. 
\begin{eqnarray*}
(\frac{1}{n_r}+\frac{1}{n_s})\sum_{i\in[N]}|x_{i}-E[x_{i}]| & \leq & (\frac{1}{n_r}+\frac{1}{n_s})\sqrt{N}\sqrt{\sum_{i\in[N]}(x_{i}-E[x_{i}])^{2}}\\
 & \leq & (\frac{1}{n_r}+\frac{1}{n_s})N\sqrt{\sigma^{2}+\epsilon_{2}}\\
 &\leq & 4M\exp(\frac{K-1}{M-1})\sqrt{\sigma^{2}+\epsilon_{2}}
\end{eqnarray*}

where the last inequality follows because from Lemma \ref{thm:num}, we have $n_s\geq \frac{16}{\epsilon^2}\ln \frac{2MK\beta(\beta+1)}{\delta}$.

\end{IEEEproof}

We now present some lemmas that are useful for proving that after clustering, the centroids are closer to the means of the distributions from which they are drawn.


\begin{lem}\label{thm:esterr}
If the separability condition \eqref{eqn:sep_clus} is satisfied, then after using Cluster algorithm, we have that for any fixed $\epsilon,\delta$ and $n_s\geq N_{\epsilon,\delta}=\lceil\frac{16}{\epsilon^2}\ln(\frac{\beta}{\delta})\rceil$, with probability greater than $1-\delta$,
\[|\hat{\mu}_s-\mu_s|= |g(S_s)-\mu_s | \leq \epsilon.\]
\end{lem}
%

\begin{IEEEproof}

From Lemma \ref{lem:approx}, after the $\alpha$ approximation algorithm, we have $\Delta_s\leq 2(\alpha+1)\frac{\phi_{\ast}}{n_s}$ and $\gamma < \frac{2(\alpha+1)}{c}$. If we want $\gamma\leq \frac{1}{8}$ which gives $a<\frac{c}{16}-1$.  From Lemma \ref{lem:rho}, $\rho_{in}^{s}+\rho_{out}^{s}\leq \frac{8}{c} $ which we need to be less than $\frac{1}{2}$ this giving us $c>16$. From this and Lemma \ref{lem:ineq}, the conditions for Lemma \ref{lem:mean} are satisfied and $\gamma <\frac{1}{8}$. Thus we have, 
\[|g(S_s)-\mu_s |\leq 2(1-\rho_{out}^{s})|g(S_s\cap {\cal T}_s)-\mu_s |+4\sum_{r\neq s:\rho_{in}^{s}(r)\neq 0}\rho_{in}^{s}(r)|g(S_s\cap {\cal T}_r)-\mu_r |.\]

For each $r\in [\beta]$, $S_{s}\cap T_{r}$ denotes independently drawn bounded random variables from reward  distribution with mean $\mu_r$, we use Hoeffding's lemma.
\begin{eqnarray*}
\Pr(\exists r\text{ s. t }|g(S_{s}\cap T_{r})-\mu_{r}|  \geq  \epsilon)&\leq&\sum_{r\in\beta}\Pr(|g(S_{s}\cap T_{r})-\mu_{r}|\geq\epsilon)\\
 & \leq_{(a)} & \exp(-2n_{s}(1-\rho_{out}^{s})(\frac{\epsilon}{4})^{2})+\sum_{r\neq s:\rho_{in}^{s}(r)\neq 0}\exp(-2n_{s}\rho_{in}^{s}(r)(\frac{\epsilon}{4})^{2})\\
 & \leq_{(b)} & \exp(-2n_{s}\rho_{in}^{s}(\frac{\epsilon}{4})^{2})+\sum_{r\neq s:\rho_{in}^{s}(r)\neq 0}\exp(-2n_{s}c_{1}(\frac{\epsilon}{4})^{2})\\
 & \leq_{(c)} & \beta\exp(-2n_{s}c_{1}(\frac{\epsilon}{4})^{2})\leq\delta,
\end{eqnarray*} 
where $c_1 = \min_{r,s}{\rho_{in}^{s}(r):\rho_{in}^{s}(r)\neq 0}$. Inequality $(a)$ follows from Hoeffding's lemma, inequality $(b)$ from $1-\rho_{out}^{s}\geq \rho_{in}^{s}$ and inequality $(c)$ from the definition of $c_1$. 

For $\beta\exp(-2n_{s}c_{1}(\frac{\epsilon}{4})^{2}))\leq\delta$, we need $n_s\geq \frac{8}{c_1\epsilon^2}\ln(\frac{\beta}{\delta}).$
Since $c_1<\frac{1}{2}$, we have 
\[n_s\geq \frac{16}{\epsilon^2}\ln(\frac{\beta}{\delta}).\]

Thus, with probability greater than $1-\delta$, we have 
\[|g(S_s)-\mu_s |\leq 2 \frac{\epsilon}{4} +4 \sum_{r\neq s:\rho_{in}^{s}(r)\neq 0}\rho_{in}^{s}(r)\frac{\epsilon}{4}\leq   \frac{\epsilon}{2} + \rho_{in}^{s}\epsilon\leq\epsilon.  \]

\end{IEEEproof}

\begin{lem}\label{lem:approx}
An $\alpha$ approximation algorithm returns the set of centroids $\{\nu_1,\dots,\nu_\beta\}$ where $C(x)$ returns the centroid of the cluster to which $x$ belongs. We have $\forall {\cal T}_s$ $\exists \nu_s$ such that $|\nu_s-\mu_s|\leq2(\alpha+1)\frac{\phi_{\ast}}{n_s}$ and $\gamma <\frac{2(\alpha+1)}{c}$.
\end{lem}

\begin{IEEEproof}

We first show that $\forall s$, $\Delta_s\leq(\alpha+1)\frac{\phi_{T}}{n_s}$ where $\phi_T=\sum_{s=1}^{\beta}\sum_{x\in {\cal T}_s}|x-g({\cal T}_s)|$. 
Assume the contrary that for some ${\cal T}_s$,$|\nu_r-\mu_s|>(\alpha+1)\frac{\phi_{T}}{n_s}$ $\forall r\in [\beta]$.
\begin{eqnarray*}
 \sum_{x\in T_{s}}|x-C(x)| & \geq & \sum_{x\in T_{s}}|C(x)-g(T_{s})|-|x-g(T_{s})|\\
 & > & |T_{s}|\frac{(\alpha+1)\phi_{T}}{|Ts|}-\sum_{x\in T_{s}}|x-g(T_{s})|\\
 & \geq & (\alpha+1)\phi_{T}-\phi_{T}=a\phi_{T},
\end{eqnarray*}
which is a contradiction.
We now show that $\phi_{T}\leq 2\phi_{\ast}$ which proves that $\Delta_s\leq 2(\alpha+1)\frac{\phi_{\ast}}{n_s}$.
\begin{eqnarray*}
\phi_{T} & = & \sum_{s=1}^{\beta}\sum_{x\in T_{s}}|x-g(T_{s})|\\
 & \leq & \sum_{s=1}^{\beta}\sum_{x\in T_{s}}|g(T_{s})-\mu_{s}|+\sum_{s=1}^{\beta}\sum_{x\in T_{s}}|x-\mu_{s}|\\
 & = & \sum_{s=1}^{\beta}|T_{s}||g(T_{s})-\mu_{s}|+\phi_{\ast}\\
 & = & \sum_{s=1}^{\beta}|\sum_{x\in Ts}x-\mu_{s}|+\phi_{\ast}\\
 & \leq & \sum_{s=1}^{\beta}\sum_{x\in Ts}|x-\mu_{s}|+\phi_{\ast}\\
 & = & 2\phi_{\ast}.
\end{eqnarray*}
Now we show that $\gamma \leq \frac{2(\alpha+1)}{c}$. For any $s,r$,
\[ \frac{2(\alpha+1)}{c}|\mu_r-\mu_s|\geq \frac{2(\alpha+1)}{c}c\phi_{\ast}(\frac{1}{n_s}+\frac{1}{n_r})\geq \Delta_s. \]
Since this is true for all $r,s$, we have 
\[\gamma \leq \frac{2(\alpha+1)}{c}.\]

\end{IEEEproof}

\begin{lem}\label{lem:ineq}
If $\gamma<\frac{1}{4}$, the following results hold $\forall x\in S_r$,
\begin{enumerate}
\item  $|x-\mu_s|\geq (\frac{1}{2}-2\gamma)|\mu_r-\mu_s|,\quad \forall s\neq r$.
\item $|x-\mu_r|\leq \frac{1}{1-4\gamma}|x-\mu_s|.$
\end{enumerate}
\end{lem}

\begin{IEEEproof}
(1) 
\begin{eqnarray*}
|\nu_{r}-\nu_{s}| & = & |\nu_{r}-\mu_{r}+\mu_{r}-\mu_{s}+\mu_{s}-\nu_{s}|\\
 & \geq & |\mu_{r}-\mu_{s}|-|\nu_{r}-\mu_{r}|-|\mu_{s}-\nu_{s}|\\
 & \geq & (1-2\gamma)|\mu_{r}-\mu_{s}|,
\end{eqnarray*}
where the last inequality follows from the definition of $\gamma$.
\begin{eqnarray*}
|x-\mu_{s}| & \geq & |x-\nu_{s}|-|\mu_{s}-\nu_{s}|\\
 & \geq & \frac{1}{2}|\nu_{r}-\nu_{s}|-|\mu_{s}-\nu_{s}|\\
 & \geq & (\frac{1}{2}-\gamma)|\mu_{r}-\mu_{s}|-|\mu_{s}-\nu_{s}|\\
 & \geq & (\frac{1}{2}-\gamma)|\mu_{r}-\mu_{s}|-\gamma|\mu_{r}-\mu_{s}|\\
 & = & (\frac{1}{2}-2\gamma)|\mu_{r}-\mu_{s}|,
\end{eqnarray*}
where the second inequality follows from $x\in S_r$ and the
last from the definition of $\gamma$.

(2)\begin{eqnarray*}
|x-\mu_{r}| & \leq & |\mu_{r}-\nu_{r}|+|x-\nu_{r}|\\
 & \leq & |\mu_{r}-\nu_{r}|+|x-\nu_{s}|\\
 & \leq & |\mu_{r}-\nu_{r}|+|x-\mu_{s}|+|\mu_{s}-\nu_{s}|.
\end{eqnarray*}
Note that the first statement with the definition of $\gamma$ also
implies for $l=r,s$ 
\[
\frac{1-4\gamma}{2\gamma}|\mu_{l}-\nu_{l}|\leq|x-\mu_{s}|,
\]
which gives us
\begin{eqnarray*}
|x-\mu_{r}| & \leq & (1+\frac{4\gamma}{1-4\gamma})|x-\mu_{s}|\\
 & = & \frac{1}{1-4\gamma}|x-\mu_{s}|.
\end{eqnarray*}

\end{IEEEproof}

\begin{lem}\label{lem:rho}
If $\gamma<\frac{1}{4}$ and $|\mu_r-\mu_s|\geq c\frac{\phi_{\ast}}{n_s}$, we have $\rho_{in}^{s}\leq \frac{2}{(1-4\gamma)c}$ and $\rho_{out}^{s}\leq \frac{2}{(1-4\gamma)c}$.
\end{lem}

\begin{IEEEproof}
From the separability condition \eqref{eqn:sep_clus}, we have $|\mu_r-\mu_s|\geq c\frac{\phi_{\ast}}{n_s}$.
\begin{eqnarray*}
n_{s}\rho_{out}^{s}(\frac{1}{2}-2\gamma)c\frac{\phi_{\ast}}{n_{s}} & \leq & \sum_{r\neq s}|T_{s}\cap S_{r}|(\frac{1}{2}-2\gamma)|\mu_{s}-\mu_{r}|\\
 & \leq & \sum_{r\neq s}\sum_{x_{i}\in T_{s}\cap S_{r}}(\frac{1}{2}-2\gamma)|\mu_{s}-\mu_{r}|\\
 & \leq & \sum_{r\neq s}\sum_{x_{i}\in T_{s}\cap S_{r}}|x_{i}-\mu_{s}|\\
 & \leq & \phi_{\ast},
\end{eqnarray*}
where the first and second inequalities follow from the separability
condition and Lemma \ref{lem:ineq} respectively. This gives us $\rho_{out}^{s}\leq\frac{2}{(1-4\gamma)c}$
and similarly we also have $\rho_{in}^{s}\leq\frac{2}{(1-4\gamma)c}$.
\end{IEEEproof}

\begin{lem}\label{lem:mean}
If $(a)\rho_{in}^{s}+\rho_{out}^{s}<\frac{1}{2}$ and $(b)|g(S_s\cap {\cal T}_r)-\mu_r |\geq (1-4\gamma)|g(S_s\cap {\cal T}_r)-\mu_s |$ we have,
\[|g(S_s)-\mu_s |\leq 2(1-\rho_{out}^{s})|g(S_s\cap {\cal T}_s)-\mu_s |+\frac{2}{1-4\gamma}\sum_{r\neq s}\rho_{in}^{s}(r)|g(S_s\cup {\cal T}_r)-\mu_r |.\]
\end{lem}

\begin{IEEEproof}
$|g(S_{s})-\mu_{s}|$
\begin{eqnarray*}
 & = & |\frac{|S_{s}\cap {\cal T}_{s}|g(S_{s}\cap {\cal T}_{s})+\sum_{r\neq s}|S_{s}\cap {\cal T}_{r}|g(S_{s}\cap {\cal T}_{r})}{|S_{s}|}-\mu_{s}|\\
 & = & \frac{|n_{s}(1-\rho_{out}^{s})(g(S_{s}\cap {\cal T}_{s})-\mu_{s})+\sum_{r\neq s}n_{s}\rho_{in}^{s}(r)(g(S_{s}\cap {\cal T}_{r})-\mu_{s})}{|S_{s}|}|\\
 & \underset{(a)}{\leq} & 2(1-\rho_{out}^{s})|(g(S_{s}\cap {\cal T}_{s})-\mu_{s})|+2\sum_{r\neq s}n_{s}\rho_{in}^{s}(r)|(g(S_{s}\cap {\cal T}_{r})-\mu_{s})|\\
 & \leq & 2[(1-\rho_{out}^{s})|(g(S_{s}\cap {\cal T}_{s})-\mu_{s})|+\sum_{r\neq s}n_{s}\rho_{in}^{s}(r)|(g(S_{s}\cap {\cal T}_{r})-\mu_{s})|]\\
 & \leq_{(b)} & 2(1-\rho_{out}^{s})|(g(S_{s}\cap {\cal T}_{s})-\mu_{s})|+\frac{2}{1-4\gamma}\sum_{r\neq s}n_{s}\rho_{in}^{s}(r)|(g(S_{s}\cap {\cal T}_{r})-\mu_{r})|.
\end{eqnarray*}

\end{IEEEproof}

\end{document}